\RequirePackage{fix-cm}
\RequirePackage{
amsmath,amsfonts,amssymb,bm,upgreek,multicol,mathrsfs,pifont,color,fancyhdr,amscd,latexsym
}
\RequirePackage{graphicx}
\RequirePackage{multirow,tabularx,url,float,booktabs}
\RequirePackage{ifpdf}
\RequirePackage{fancyhdr,lastpage}
\RequirePackage{supertabular}
\RequirePackage[pdfstartview=FitH,colorlinks,breaklinks,linkcolor=black,citecolor=black,filecolor=black,urlcolor=black,hyperindex,CJKbookmarks]{hyperref}
\RequirePackage{silence}
\documentclass[twocolumn]{svjour3}          
\smartqed  
\usepackage{graphicx,subfigure}
\usepackage[linesnumbered, ruled, vlined]{algorithm2e}
\usepackage{diagbox}
\usepackage{multirow}
\usepackage{subfigure}
\graphicspath{{./}}
\journalname{CGI2024} 
\newcommand*{\bigE}{\scalebox{1.5}{\ensuremath{\mathcal{\varepsilon}}}}
\newcommand*{\smallbigE}{\scalebox{1.0}{\ensuremath{\mathcal{\varepsilon}}}}

\newcommand*{\smallbigX}{\scalebox{0.6}{\ensuremath{\mathcal{\mathrm{X}}}}}
\begin{document}
\begin{sloppypar}
\title{GPN: \underline{G}enerative \underline{P}oint-based \underline{N}eRF}
\subtitle{}
\author{Haipeng Wang
}

\authorrunning{Haipeng Wang.}
\institute{
	Haipeng Wang \at College of Mechanical Engineering, Zhejiang Sci-Tech University,\\
	\email{201810501019@mails.zstu.edu.cn}
}
\date{ }

\maketitle

\begin{abstract}

Scanning real-life scenes with modern registration devices typically gives incomplete point cloud representations, primarily due to the limitations of partial scanning, 3D occlusions, and dynamic light conditions. Recent works on processing incomplete point clouds have always focused on point cloud completion. However, these approaches do not ensure consistency between the completed point cloud and the captured images regarding color and geometry. We propose using Generative Point-based NeRF (GPN) to reconstruct and repair a partial cloud by fully utilizing the scanning images and the corresponding reconstructed cloud. The repaired point cloud can achieve multi-view consistency with the captured images at high spatial resolution. For the finetunes of a single scene, we optimize the global latent condition by incorporating an Auto-Decoder architecture while retaining multi-view consistency. As a result, the generated point clouds are smooth, plausible, and geometrically consistent with the partial scanning images. Extensive experiments on ShapeNet demonstrate that our works achieve competitive performances to the other state-of-the-art point cloud-based neural scene rendering and editing performances.

\end{abstract}

\section{Introduction}\label{IntroSec}
In recent years, we have seen impressive progress in computer graphics and vision with the advancements in 3D reconstruction and differentiable rendering. Thanks to affordable RGB / RGB-D cameras, commercial 3D reconstruction applications have become more accessible. However, the scanning process of a scene may be incomplete for various reasons. As the scene's complexity increases, current methods face the main obstacles in solving this problem, such as limited sensor resolution, difficulty covering all scanning positions, and 3D occlusions, resulting in incomplete point clouds. Some recent works have proposed methods that focus on generating 3D representations for novel scenes, like some Text-to-2D diffusion models with 3D Gaussian splatting~\cite{3DGaussianSplatting} representation~\cite{GaussianDiffusion}~\cite{Dreamgaussian}. However, it is essential to note that these methods are only suitable for creating small-scale 3D entertainment assets and do not apply to real-life commercial industries. Moreover, while Gaussian Splatting offers a similar rendering quality with faster training and inference, it is challenging to condition since it usually requires circa hundred thousand Gaussian components~\cite{VDGS}. High GPU memory requirements make extending these diffusion-based generative models to 3D representation completion work in industrial scenarios challenging. 
\par
We proposed a lightweight, generalizable point-based NeRF framework for partial scanning data processing. This framework enables generative reconstruction of the partial point cloud, especially for the completion task, while simultaneously following the partial scene geometry and the scanning images to keep multi-view consistency. Compared to a series of generative frameworks based on Gaussian splatting and diffusion modeling, our framework used the hypernetwork paradigm-based VAE architecture that could generate novel scenes with high-quality reconstruction and low GPU memory. Additionally, a pre-trained GPN could reconstruct an unseen scene without images, only the cloud as input. Meanwhile, generalizable NeRF, such as CodeNeRF~\cite{CodeNeRF}, GRF~\cite{GRF}, GRAF~\cite{GRAF}, GIRAFEE~\cite{GIRAFFE} needs at least one image as input.
\par
According to the reconstructed point cloud's completeness, we propose a basic ``Generation Framework'' dedicated to rendering and reconstructing complete point clouds and a ``Completion Framework'' dedicated to repairing and complementing incomplete point clouds. Inspired by HyperCloud~\cite{HyperCloud}, the ``Generation Framework'' uses a single VAE generative architecture for generalization training. We embody the hypernetwork paradigm as we take a 3D colored point cloud as an input and return the parameters of the NeRF network. The ``Generation Framework'' allows for conditioning NeRF, and whose parameters can be interpreted as a continuous parameterization of a 3D object space. Inspired by Hyperpocket~\cite{HyperPocket}, the ``Completion Framework'' uses two encoders to produce separate latent representations for the existing part of point clouds and its missing part. Both representations are concatenated and passed through a decoder, which outputs the weights of NeRF used to reconstruct and render a complete point cloud.
\par
During inference, the encoder predicts an initial latent code and generates a unique NeRF model for the input cloud. To match the reconstructed cloud and the existing scanning images, we perform latent-based conditional optimization to refine the predicted NeRF model.
\par
In summary, our work makes the following contributions:
\begin{itemize}
	\item
	We propose a lightweight, generalizable, point-based NeRF framework that combines a hypernetwork with the point-based NeRF model and continuously represents 3D objects as NeRF parameters.
	\item 
	We introduce the ``Generation Framework'' that could reconstruct the input complete cloud into a high spatial resolution surface directly, and the ``Completion Framework'' could complement the incomplete point cloud with colors. 
	\item
	We propose the first colored point cloud completion process based on Point-based NeRF, which allows us to complete the partial scanning data that follow the input scene geometry while simultaneously adapting the generations to the scanning images.
\end{itemize}
\section{Related Work}\label{RWSec}
\textbf{NeRF-based Generative Framework.} 
There are mainly four architectures for the NeRF-based generalization framework: GAN, auto-decoder, and auto-encoder, diffusion model. GRAF~\cite{GRAF}, GRF~\cite{GRF}, and GIRAFFE~\cite{GIRAFFE} introduce the idea of conditional radiance fields into the GAN-based generative adversarial network framework to convert it into a 3D aware generative model. HyperNeRFGAN~\cite{HyperNeRFGAN} combines the Hypernetwork paradigm and GAN-based structure to produce 3D objects represented by NeRF. Points2NeRF~\cite{Points2NeRF} leverages auto-encoder architecture and the hypernetwork training paradigm to produce a continuous representation of objects. To learn category-level priors, methods like CodeNeRF~\cite{CodeNeRF} and LOLNeRF~\cite{LOLNeRF} use an auto-decoder-based conditional NeRF on instance vectors, where different latent result in different NeRFs. MVDD~\cite{MVDD} uses multi-view depth representation with diffusion models to generate 3D consistent multi-view depth maps for shape generation and completion. Latent-NeRF~\cite{Latent-NeRF} enables the training of DreamFusion~\cite{DreamFusion} with higher-resolution images by optimizing the NeRF with diffusion model features instead of RGB colors. SSDNeRF~\cite{SSDNeRF} models the generative prior of scene latent codes with a 3D latent diffusion model.
\par
\noindent \textbf{Point-based NeRF.} Point2Pix~\cite{Point2Pix} predicts corresponding 3D attributes in the nearby area of existing points through a point encoder that can extract multi-scale point features and a fusion-based decoder to synthesize realistic images. TriVol ~\cite{TriVol} introduces a 3DUNet-based feature extractor and aggregator to enhance the rendering ability of NeRF. Point-NeRF~\cite{PointNeRF} renders and completes the missing parts of the reconstructed point cloud through a pre-trained MVSNet~\cite{MVSNet}, finding the missing depths from existing captured images. Point-NeRF++~\cite{PointNeRF++} introduces an effective multi-scale representation for point cloud-based rendering. Point-SLAM~\cite{Point-NeRF-SLAM} introduces a neural point-based NeRF scene representation and can be effectively used for mapping and tracking in RGB-D SLAM. Points2NeRF~\cite{Points2NeRF} adapts the hypernetwork framework to the NeRF architecture and allows the production of NeRF from the point cloud.
\par
\noindent \textbf{Hypernetwork paradigm-based generative model.} HyperNet~\cite{HyperNetworks} dynamically generates the weights of a model with variable architecture. HyperDiffusion~\cite{HyperDiffusion} addresses the unconditional generative modeling of NeRF by operating directly on MLP weights. MetaDiff~\cite{MetaDiff}, NeuralNetworkDiffusion~\cite{NeuralNetworkDiffusion} introduce a meta-learning method for learning the diffusion model parameters. HyperCloud~\cite{HyperCloud}, HyperPocket~\cite{HyperPocket} is a generative auto-encoder-based model that maps the probability distributions to 3D models with MLPs trained by a hypernetwork. Hyp-NeRF~\cite{HyP-NeRF} uses hypernetwork to learn prior weights of NeRF MLPs, and tightly couples text and NeRF priors to generate and edit NeRFs based on text inputs.
\begin{figure*}[htbp]
	\centering
	\includegraphics[scale=0.06]{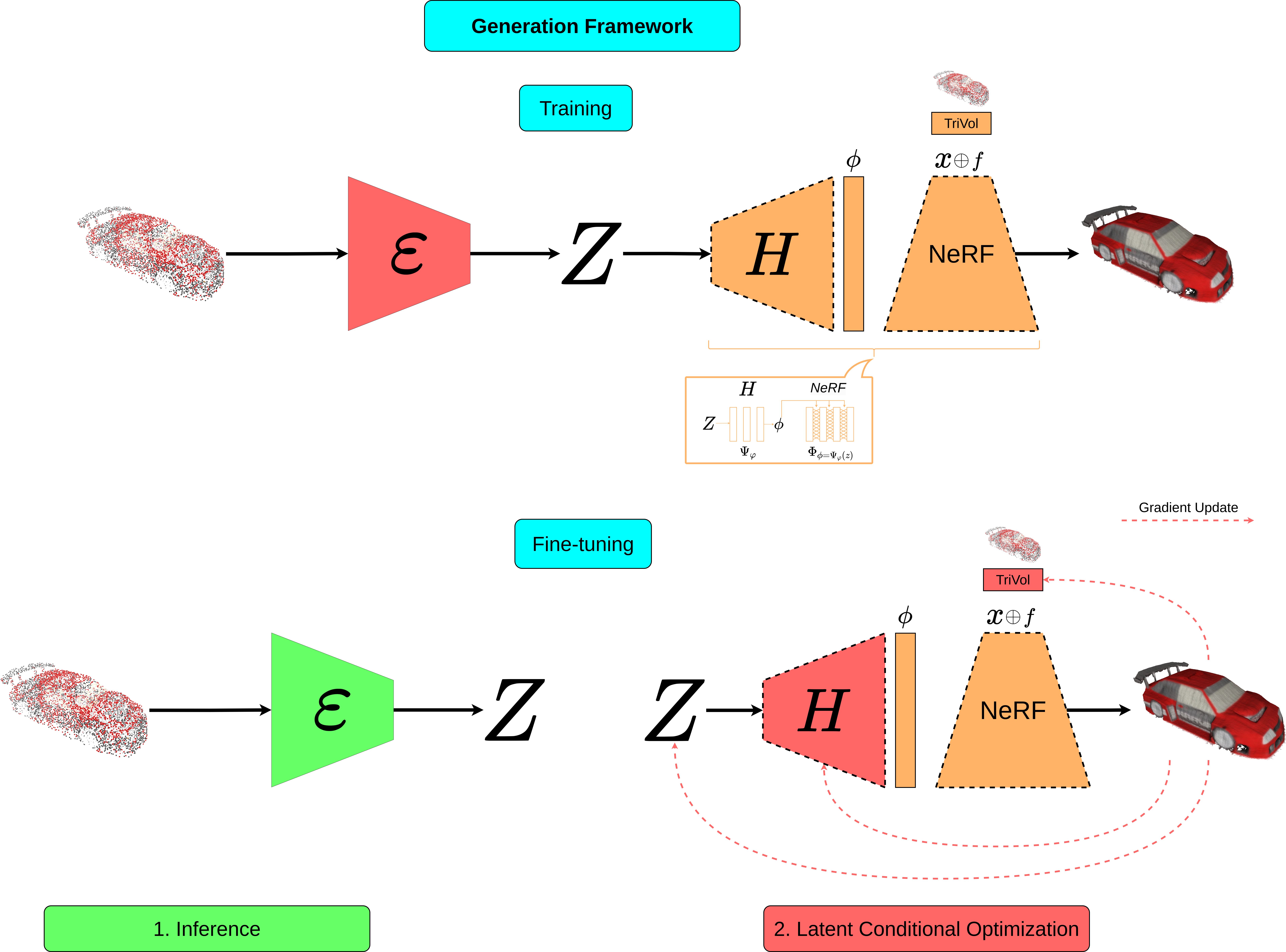}
	\caption{Flow diagram of the Generation Framework.}
	\label{GenerationFrameworkFig}
\end{figure*}
\section{Overview}\label{OverviewSec}
We present the Generative Point-based NeRF, which can handle incomplete Point Cloud generated from incomplete scanning. Our framework, GPN, employs a hypernetwork paradigm-based VAE to approximate the MLP weights of the NeRF neural network. Additionally, we introduce an auto-decoder-based finetune strategy to enhance the reconstruction quality while maintaining the multiview consistency of the generated NeRF. We propose two frameworks, the ``Generation Framework'' and the ``Completion Framework'', for reconstructing and repairing incomplete point clouds. Sections \ref{GenerationFrameworkSec} and \ref{CompletionFrameworkSec} provide a detailed explanation of these frameworks.
\section{Generation Framework}\label{GenerationFrameworkSec}
We propose the basic ``GPN: Generation Framework'', which leverages a hypernetwork architecture that takes a 3D point cloud as an input while returning the parameters of the NeRF network. Since The hypernetwork generates the parameters of NeRF, the reconstructed implicit NeRF scene can differ based on different point clouds. This results in a continuous parameterization of the object's surface and the NeRF implicit representation of the input point cloud. An overview of the ``Generation Framework'' is illustrated in Fig.\ref{GenerationFrameworkFig}.
\par
\subsection{Hypernetwork paradigm-based VAE} \label{VAESec}
Our approach is constructing a VAE architecture, which takes as an input a colored point cloud and generates the weights of the NeRF network. We use a hypernetwork paradigm to aggregate basic shape and color information from colored point clouds. This paradigm could reduce the number of trainable NeRF network parameters by designing a Hypernetwork with fewer parameters. Additionally, in the context of 3D objects, various methods~\cite{HyperCloud}~\cite{HyperPocket}~\cite{HyperNeRFGAN} ~\cite{HyperDiffusion} make use of a hypernetwork to produce a continuous representation of objects, even a continuous implicit NeRF representation~\cite{MetaDiff}. The Hypernetwork paradigm-based generative model uses different weights for each 3D object. Roughly speaking, instead of producing point clouds, we would like to produce many NeRF implicit scenes (a specific NeRF for each point cloud) that model them. 
\par
The architecture of VAE consists of: an encoder $\bigE$, which is a PointNet\cite{PointNet}-like network that transports the colored point cloud to lower-dimensional latent space $\mathrm{Z}\subseteq \Re^{\mathrm{D}}$, and a decoder $\mathrm{\Psi}_{\psi}$, which transfers latent space to the vector of weights $\phi$ for the NeRF $\mathrm{\Phi}_{\phi}$. We present the parameterization of the 3D objects as a function $\mathrm{\Phi}_\phi : \Re^3 \to \Re^4$, which given location $(x,y,z)$ returns a color ${\bf c} = (r, g, b)$ and volume density $\sigma$. We differ from the standard NeRF in one more aspect, as we do not use the viewing direction. That is because the images used for training do not have view-dependent features like reflections. 
\par 
The NeRF $\mathrm{\Phi}_\phi\left ( x,y,z \right ) $ is not trained directly. We use a hypernetwork $\mathrm{\Psi}_\psi: \Re^6\supset \mathrm {X}   \to \phi$ which for a point cloud $\mathrm{X} \subset \Re^6$ returns weights $\phi$ to the corresponding NeRF $\mathrm{\Phi}_\phi$. Thus, a point cloud $\mathrm{X}$ is represented by $\mathrm{\Phi}\left ( \mathrm{X},\phi \right ) = \mathrm{\Phi}\left ( \mathrm{X},\mathrm{\Psi}_{\psi}\left ( \mathrm{Z} \right )  \right ) = \mathrm{\Phi}\left ( \mathrm{X},\mathrm{\Psi}_{\psi}\left ( \bigE \left ( \mathrm{X} \right )  \right )  \right )$.
\par
\begin{figure*}[htbp]
	\centering
	\includegraphics[scale=0.06]{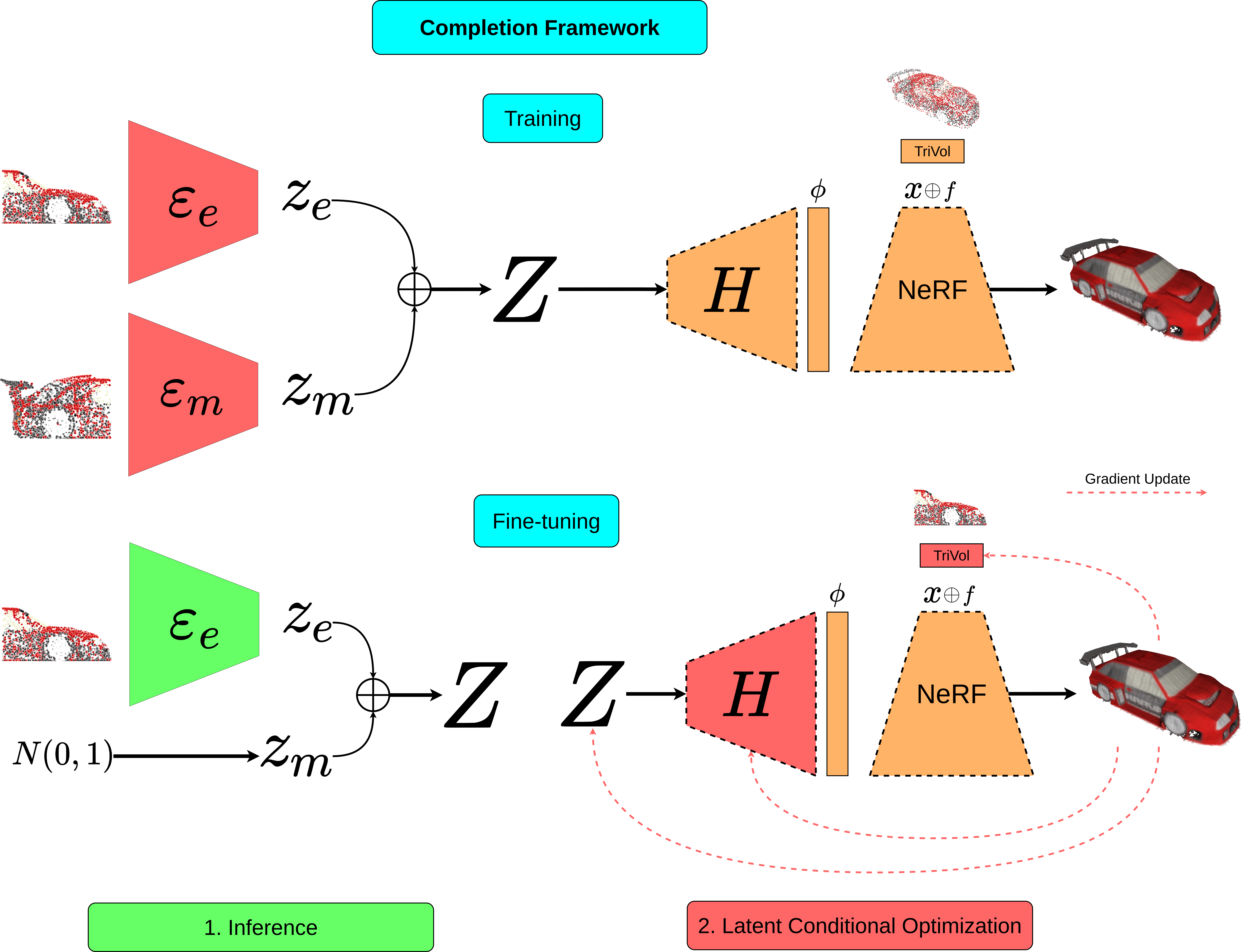}
	\caption{Flow diagram of the Completion Framework.}
	\label{CompletionFrameworkFig}
\end{figure*}
\subsection{Point-based NeRF Rendering}\label{PNRSec}
Recent works on point-based NeRF rendering have been primarily focused on how to sample effectively based on the input point cloud. There are mainly two sample methods: point-based~\cite{BoostingPointNeRF}~\cite{PointNeRF}~\cite{PointNeRF++} and voxel-based~\cite{Point2Pix}~\cite{TriVol}. The difference is that the point-based sampling method is equivalent to projecting the point cloud onto the novel view directly in theory. The geometry and color of the whole scene are determined by the points uniquely and, most importantly, could not be trained for generalization. However, the voxel-based sampling method is learning-based and could approximate the scene's occupancy grid~\cite{NeRFAcc} by learning the NeRF network parameters. Unlike vanilla NeRF~\cite{NeRF}, the voxel-based sampling-based rendering could render a novel view quickly, depending on the trained scene-specific NeRF parameters. Therefore, we adopt the voxel-based sampling-based rendering method for generalization training.
\par
Extended from the TriVol~\cite{TriVol} rendering method, we enhance the rendering results of the NeRF model through a volume-based feature extractor and aggregator. We voxelize the original point cloud into the 3D representation as the TriVol proposed, then utilize three 3D UNet networks to decode it into the feature representation. Besides, we adopt the NeRFAcc~\cite{NeRFAcc}, which incorporates occupancy grid-based sampling methods from Instant-NGP~\cite{Instant-NGP}. Finally, under the occupancy grid-based sampling method, combined with the high-quality feature extractor and aggregator, our point-based rendering method could return a mesh representation of the point cloud at virtually no cost in the quality of reconstructions and significantly decrease the generalization training time and GPU memory.
\subsection{Training}\label{GenTrainingSec}
We have hypernetwork-based VAE generative architecture that use different weights $\phi$ for each point cloud $\mathrm{X}$. The NeRF $\mathrm{\Phi}_{\phi}$ network is not trained directly. To use our ``Generation Framework'', we need to train the weights $\psi$ of the Hypernetwork $\mathrm{\Psi}_{\psi}$ instead. We minimize the NeRF $L2$ rendering loss $\mathcal{L}_{\mathrm{color}}$ over the training set, supervising our rendered pixels from ray marching with the ground truth $C\left (\mathbf{r} \right )$. 
\begin{align}
\mathcal{L_{\mathrm{color}}}=\sum_{r\in R}\left \| \mathrm{\Phi}_{\phi=\mathrm{\Psi}_{\psi\left ( \smallbigE \left ( \smallbigX \right )  \right ) }}\left ( \mathbf{r}  \right )-C\left (\mathbf{r} \right )    \right \| ^2_2
\label{ColorLossFunctionEQU}
\end{align}
\par
\noindent We take a point cloud $\mathrm{X}\subset \Re ^6$ and pass it to $\mathrm{\Psi}_{\psi}$, which will return weights $\phi$ to the NeRF $\mathrm{\Phi}_{\phi}$, where $\mathrm{\Phi}_{\phi} \left( \mathbf{r} \right)$ is the predicted RGB colors for ray $\mathbf{r}$. 
\par
VAE are generative models that are capable of learning approximated data distribution by applying variational inference~\cite{VAELoss}. To ensure that the data transported to latent space $\mathcal{Z} $ are distributed according to standard normal density. We add the distance from standard multivariate normal density.
\begin{equation*}
	\mathcal{L} = \mathcal{L}_{\rm{color}} + \mathcal{D}_{\rm{KL}}\left( {\bigE \left(\mathrm{X} \right),N\left( {0,\mathrm{I}} \right)} \right)
	\label{wholeLossFunctionEQU}
\end{equation*}
where $\mathcal{D}_{\rm{KL}}$ is the Kullback-Leibler divergence.
\subsection{Fine-tuning}\label{FineTuningGen}
If we have a reconstructed point cloud of an object that has not been seen before, we can directly generate a novel NeRF implicit scene through the pre-trained ``Generation Framework''. We could obtain the voxel representation of the object using the volume density samples along rays. Then, we can reconstruct the mesh using the marching cubes method. Finally, by using pre-defined novel camera poses, we can predict the color for all vertices from the rendered nearest novel view.
\par
We can treat the fine-tuning as a per-scene optimization process and follow the training procedure as a single-scene NeRF. Fig.\ref{CompletionFrameworkFig} depicts our fine-tuning pipeline. We condition the NeRF $\mathrm{\Phi}_{\phi}$ network on the input point cloud $\mathrm{X}$ via a PointNet-like encoder $\bigE$. The encoder maps the point cloud $\mathrm{X}$ into a latent vector, $z$. Inspired by the fine-tuning process of LOLNeRF~\cite{LOLNeRF} and Code-NeRF~\cite{CodeNeRF}, we optimize the reconstructed mesh using the rendering loss $\mathcal{L}_{\rm{color}}$ supervised by the scanned color frames.
\section{Completion Framework}\label{CompletionFrameworkSec}
Many current methods~\cite{Snowflakenet}~\cite{SDS-complete} for point cloud completion rely on learning to reconstruct a synthetic object, rather than completing the unseen portions of a real-world object. To address this limitation, we propose an advanced generation framework called ``Completion Framework'' that repairs and completes reconstructed point clouds from RGB / RGB-D frames that only partially capture the object. Inspired by the Hyperpocket~\cite{HyperPocket}, the ``Completion Framework'' uses two encoders to produce separate latent representations for an object's existing and missing parts. The fine-tuning process can produce more realistic point cloud completions by enforcing a Gaussian distribution on the latent space representing the missing part and maintaining the multi-view consistency with scanned images. Please see Fig.\ref{CompletionFrameworkFig} for an overview of the whole diagram.
\par
\subsection{Multi-conditional Embedding}
\begin{figure}[htbp]
	\centering
	\includegraphics[scale=0.1]{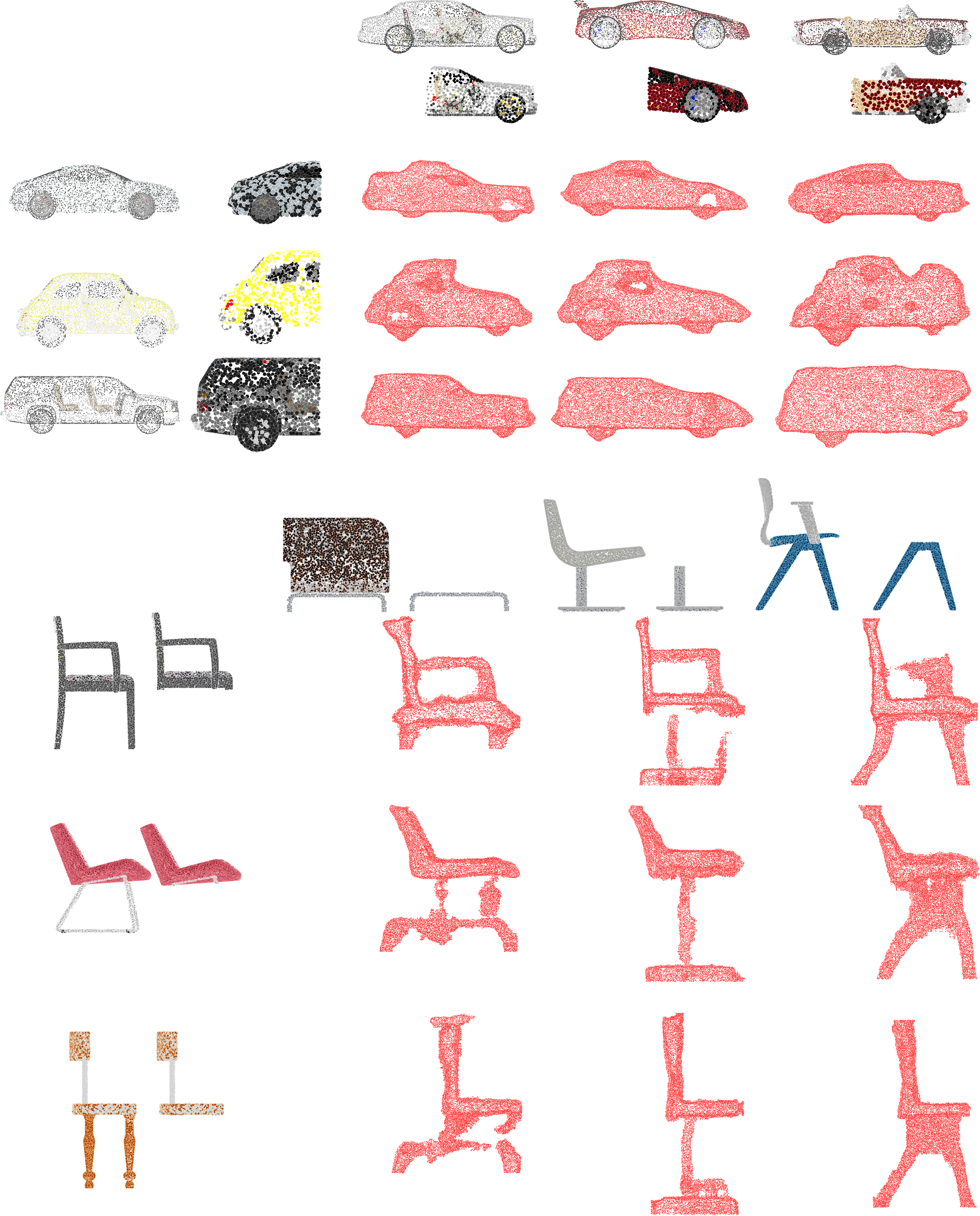}
	\caption{Novel object generation from different parts of objects. This figure shows the reconstruction of point clouds stitched together from different parts. The first row represents the point cloud parts from one object, while the first column represents parts from another. The remaining portion of the figure demonstrates the stitched reconstructions performed by the ``Completion Framework'', which inherits geometrical properties from different parts of different objects.}
	\label{NOGFig}
\end{figure}

\begin{figure}[htbp]
	\centering
	\includegraphics[scale=0.075]{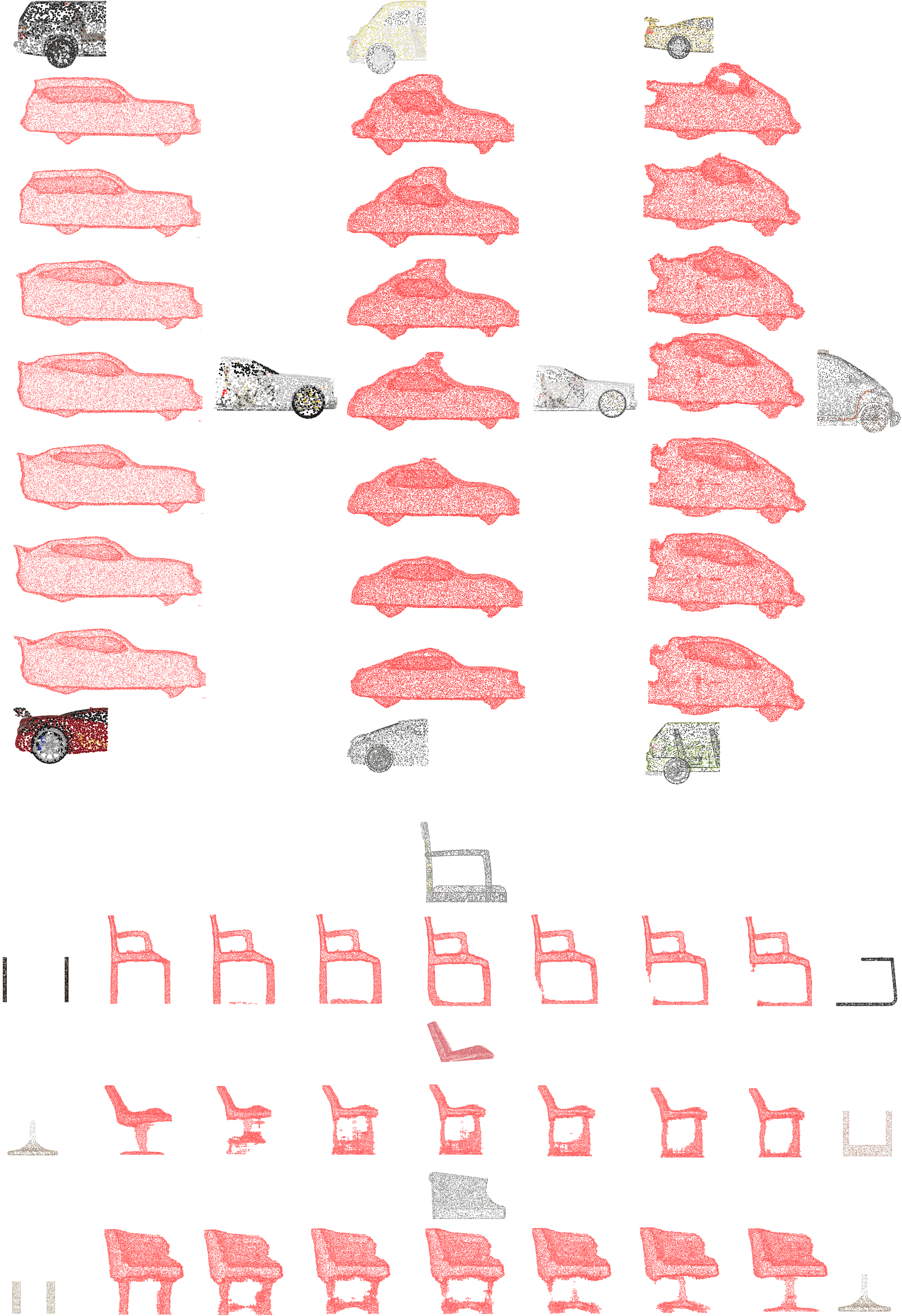}
	\caption{Interpolations between different object parts. To perform interpolation between different objects, we cut the partial cloud from one object as a constraint and then perform interpolation between two partial clouds from two different objects. We can observe that the surface points of the generated cloud perform continuous deformation between two partial clouds.}
	\label{LSIFig}
\end{figure}

We divide the input point cloud into two disjoint subsets, $P_e$ and $P_m$, representing existing and missing parts of an object, respectively. These subsets are then treated as inputs to the two independent encoders $\bigE_e$ and $\bigE_m$. The $\bigE_e$ processes $P_e$ and is used only for reconstruction. The $\bigE_m$ works with $P_m$ and renders a generative model since we force $z_m=\bigE_m \left( P_m \right)$ to follow the Gaussian prior. The two representations $z_e$ and $z_m$ are then combined and passed through a decoder, which outputs the weights of a NeRF network to produce a complete NeRF implicit scene. However, latent space encodes each part of the object separately; such a concatenation is inconsistent since we independently model $z_e$ and $z_m$. Thanks to the hypernetwork architecture, as the Fig.\ref{NOGFig} and Fig.\ref{LSIFig} shows, the NeRF network merges them into a single implicit representation, which could generate a smooth and self-consistent point cloud~\cite{HyperPocket}. 
\par
\par
\subsection{Training}
We assume that we are given a splitting of the point cloud $P$ into disjoint subsets $P_e$ and $P_m$. We use two encoders $\varepsilon_e$ and $\varepsilon_m$ for producing two disentangled yet complementary latent representations $z_e$ and $z_m$. In the training phase, slightly different from the ``Generation Framework,'' we train the VAE model to transfer two incomplete clouds into a unified and complete NeRF implicit field. Similarly, we adopt a loss function that combines the rendering loss and the KL divergence.
\begin{align}
\mathcal{L_{\mathrm{color}}}=\sum_{r\in R}\left \| \mathrm{\Phi}_{\phi=\mathrm{\Psi}_{\psi\left ( \smallbigE_e \left ( P_e \right ) \oplus  \smallbigE_m \left ( P_m \right ) \right ) }}\left ( \mathbf{r}  \right )-C\left (\mathbf{r} \right )    \right \| ^2_2
\\
\mathcal{L} = \mathcal{L}_{\rm{color}} + \mathcal{D}_{\rm{KL}}\left( {\bigE_e \left( P_e \right) \oplus \bigE_m \left ( P_m \right ),N\left( {0,\mathrm{I}} \right)} \right)
\end{align}
where $\bigE_e \left( P_e \right) \oplus \bigE_m \left ( P_m \right )$ means concatenation of two latent representations.
\subsection{Fine-tuning} \label{FineTuningCom}
\begin{figure*}[htbp]
	\centering
	\includegraphics[scale=0.15]{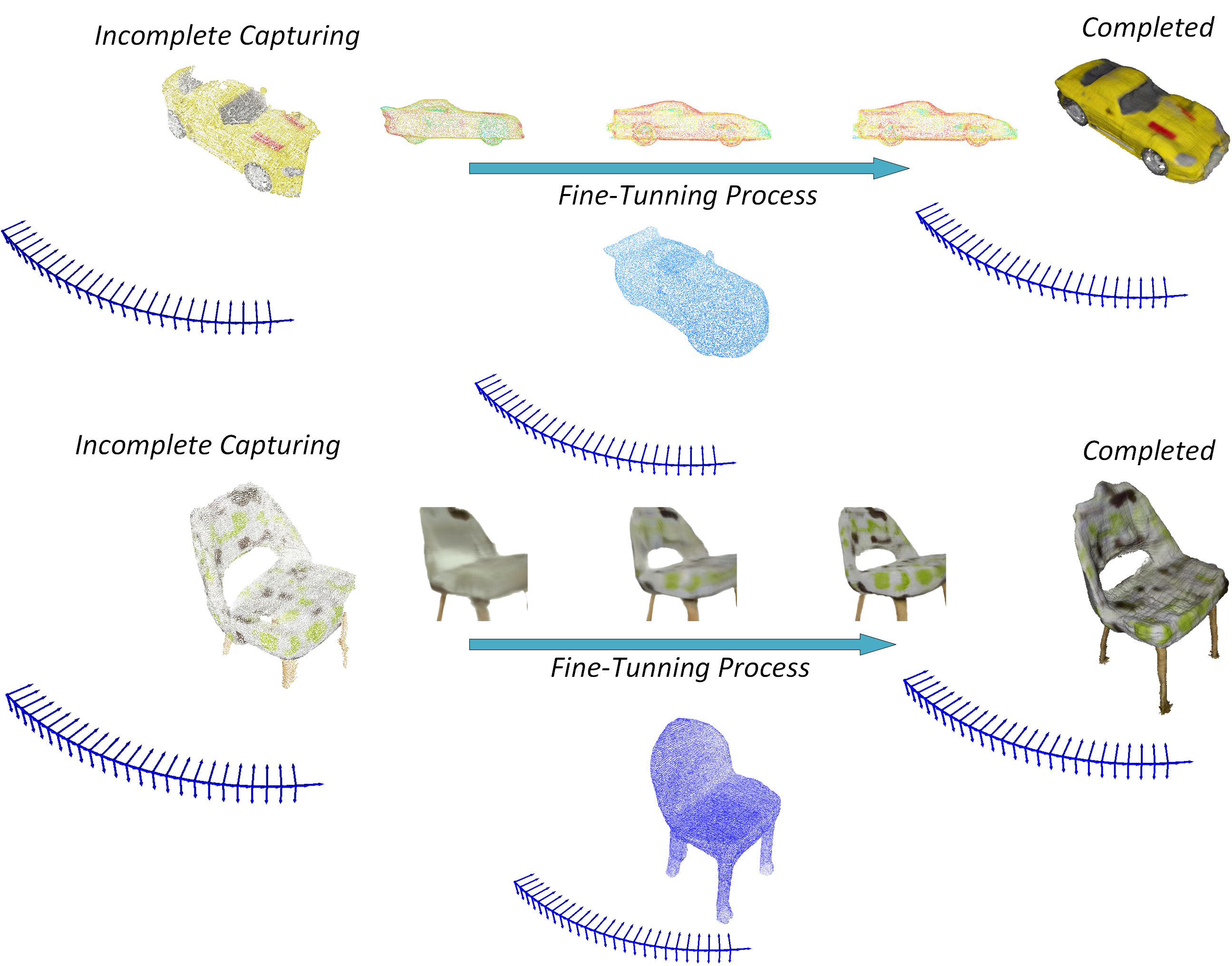}
	\caption{Completion Process. We simulate an incomplete scanning process and observe that the completed cloud geometry becomes more consistent with the captured frames, even if it does not cover the entire object. The completed surface mesh also has a more distinct texture after fine-tuning compared to the reconstructed result before fine-tuning.}
	\label{CPFig}
\end{figure*}
\par
In the fine-tuning stage, given an incomplete reconstructed point cloud and the scanned color frames that partly cover the object, the pre-trained ``Completion Framework'' could produce diversified completions of the incomplete point clouds by sampling $z\in Z_m$ from the standard Gaussian distribution: $D\left ( \bigE_e \left( P_e\right) ,z \right ) $ where $z\sim N_{Z_m}\left ( 0, I \right ) $. Therefore, using the generative aspects of the ``Completion Framework'', we can efficiently modify the produced reconstruction by changing $z\in Z_m$ to satisfy the desired constraints. Therefore, we make $z_m$ a trainable parameter and minimize the rendering loss from the scanned color frames:
\begin{align}
\mathcal{L} = \mathcal{L_{\mathrm{color}}}=\sum_{r\in R}\left \| \mathrm{\Phi}_{\phi=\mathrm{\Psi}_{\psi\left ( z_m  \right ) }}\left ( \mathbf{r}  \right )-C\left (\mathbf{r} \right )    \right \| ^2_2
\label{ColorLossFunctionComFinetunningEQU}
\end{align}
\par
Fig.\ref{CPFig} illustrates that the geometry and color of the completed point cloud can be optimized to achieve multi-view consistency with the scanned color frames.
\section{Experiments and Applications} \label{RESec}
\begin{figure*}[htbp]
	\centering
	\includegraphics[scale=0.25]{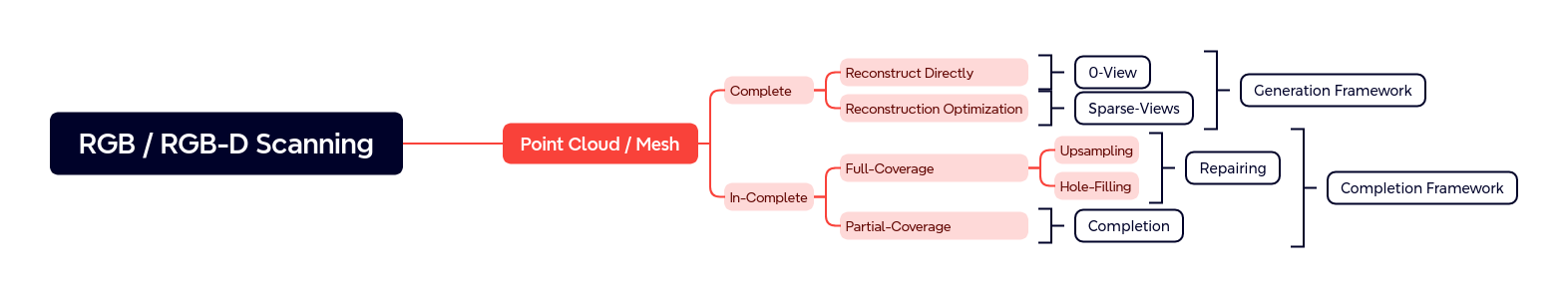}
	\caption{GPN Usage.}
	\label{GPNUsageFig}
\end{figure*}
\par
Our GPN framework primarily deals with RGB or RGB-D scanning reconstruction applications. To compare our method with other Generative Point-based NeRF methods, we will focus on specific applications as Figure.\ref{GPNUsageFig} shows.
\par
\noindent \textbf{Dataset.} 
We evaluate our framework's performance on standard shape generation benchmark categories such as cars and chairs. We split the first 90$\%$ of scenes as a training set and the rest as the testing set. The ShapeNet dataset's point clouds are obtained by 3D mesh sampling, and Blender generates the ground truth rendered images. We uniformly sample 16384 colored points on each 3D model and 100 views in a hemisphere centered on the 3D model's center with a radius of 1.5. In the Completion Framework, we randomly sample a 3D plane segment within the enclosing box of a 3D model and sample colored points into the existing and missing parts. The ShapeNet's training images have a resolution of 200 x 200.
\par
\noindent \textbf{Setup.} We set the volume resolution of TriVol to 64, and the number of layers in NeRF's MLP is 4. We use the NeRFAcc~\cite{NeRFAcc} as our rendering backend, and the view direction would not be fed into the NeRF network. For each iteration, we randomly sample 400 rays from one viewpoint. The volume resolution in NeRFAcc is 64 for ray sampling.  AdmW is adopted as the optimizer, where the learning rate is initialized as 0.001. We train models on ShapeNet using one RTX 3060Ti GPU with 8GB of GPU memory and per-scene optimization using NVIDIA 1060 GPU with only 6GB of GPU memory.
\begin{table}[htbp]
	\centering
	\resizebox{\columnwidth}{!}{%
		\begin{tabular}{l|lll}
			Point Cloud Processing Ability & Ours & TriVol\cite{TriVol} & Points2NeRF\cite{Points2NeRF} \\ \hline
			Rendering \& Reconstruction &   \multicolumn{1}{c}{$\surd$}   &    \multicolumn{1}{c}{$\surd$}    &     \multicolumn{1}{c}{$\times$}        \\
			Repairing \& Completion     &   \multicolumn{1}{c}{$\surd$}   &    \multicolumn{1}{c}{$\times$}    &    \multicolumn{1}{c}{$\surd$}         \\
			Generation \& Interpolation &   \multicolumn{1}{c}{$\surd$}   &    \multicolumn{1}{c}{$\times$}    &    \multicolumn{1}{c}{$\surd$}        
		\end{tabular}%
	}
	\caption{Generative Abality comparison among some state-of-art Generative Point-based NeRF methods.}
	\label{ComparisonOtherGPNFig}
\end{table}
\subsection{Baselines}
We compare GPN with recent generative point-based NeRF frameworks such as TriVol~\cite{TriVol} and Points2NeRF~\cite{Points2NeRF}. However, these approaches have different emphases in applications related to incompletely captured point cloud processing, and a brief comparison is shown in Table.\ref{ComparisonOtherGPNFig}. The main objective of TriVol is to achieve one-click photo-quality point cloud rendering by generalized training on a large amount of data. Since TriVol requires point clouds with as high density as possible, applying it to applications such as completion or repair is difficult. Points2NeRF learns the implicit latent representation of each 3D model in the dataset, which can be used in the point cloud-related processing tasks mentioned in this paper. However, the generated neural radiance field is obscure since its input point cloud is only a tiny part of the original point cloud. In other hands, both method could not execute the fine-tuning process, because they are not designed for per-scene optimization.
\par
\subsection{Reconstruction} \label{RESec4Rendering}
Direct reconstruction into surface mesh from point clouds is a classical problem in computer vision. Our GPN framework, based on NeRF implicit representation, could not only reconstruct directly from the point cloud but also optimize the color and geometry of the reconstructed surface from the existing captured color frames. 
\par
\noindent \textbf{0-View.} This application results mean we could reconstruct colored surface mesh from colored cloud directly based on the pre-trained GPN Generation framework.
\begin{figure}[htbp]
	\centering
	\includegraphics[scale=0.2]{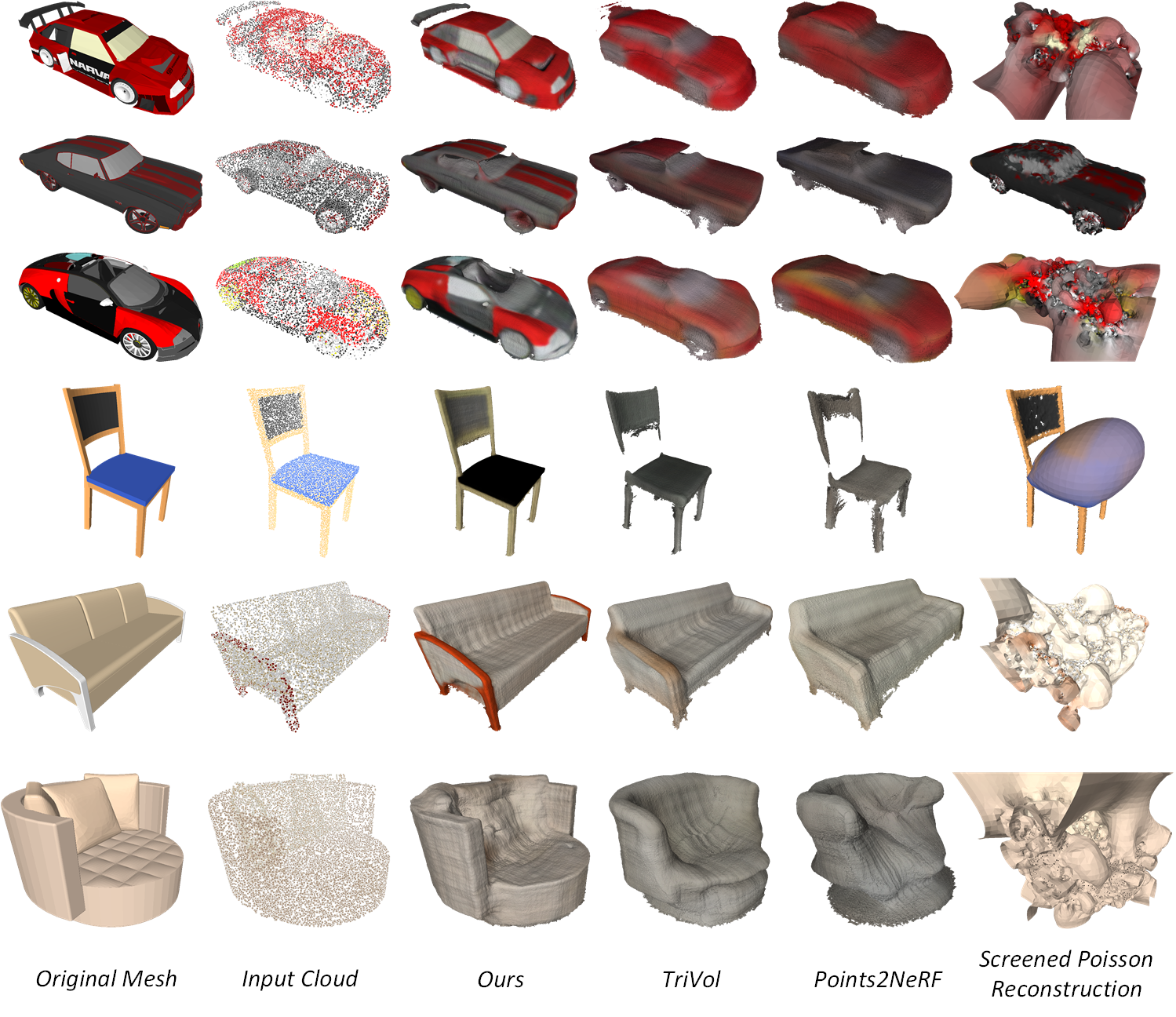}
	\caption{0-View Reconstuction. We could observe that our results perform a better reconstruction than other state-of-art point-based NeRF framework. And because of the low density of point cloud, the screened poisson reconstruction are failure in most of scenes.}
	\label{ZVFig}
\end{figure}
\par
\noindent \textbf{Sparse-Views.} This application results mean we could optimize the texture and geometry of the reconstructed surface mesh using the existing images.
\begin{figure}[htbp]
	\centering
	\includegraphics[scale=0.12]{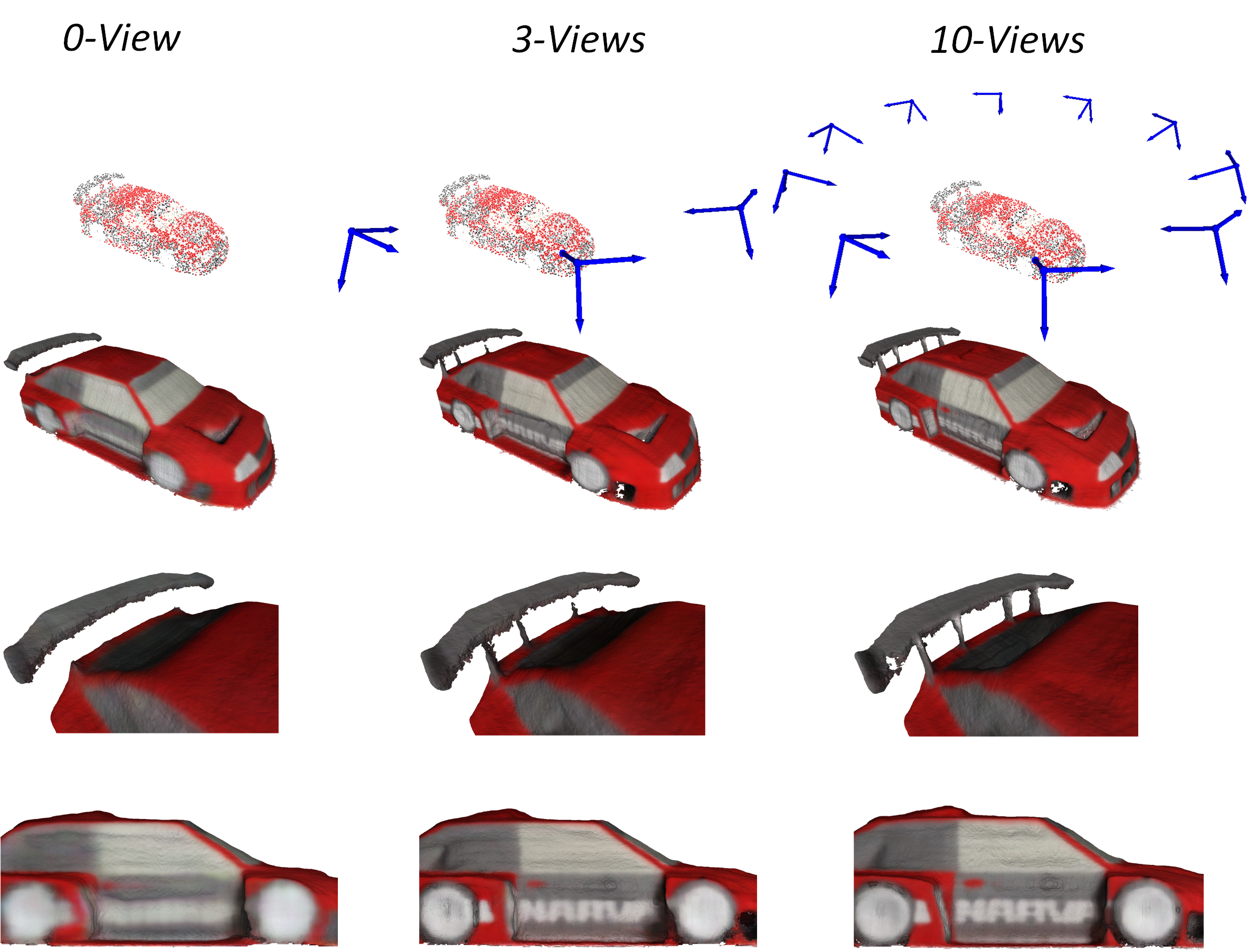}
	\caption{Sparse-View Reconstruction and Optimization. With the increasing num of existing frames, we could optimize the geometry and texture of the reconstructed surface based on the fine-tuning on the existing frames.}
	\label{SVFig}
\end{figure}
\par
\noindent \textbf{Complete Views.} It should be noted that the complete views mean we have complete reconstructed cloud and full coverage capturing images. A comparion Table.\ref{CompleteViewsTable} shows that our method could render a high-quality photo-realistic view within low-density point cloud as input while other method should input high-density point cloud. We adopt the metrics of PSNR, SSIM, and LPIPS for evaluation.
\begin{table}[htbp]
	\centering
	\resizebox{\columnwidth}{!}{%
		\begin{tabular}{l|cccc}
			Rendering & \multicolumn{1}{l}{$\rm{Ours}_{\rm{2K}}$} & \multicolumn{1}{l}{$\rm{Ours}_{\rm{10K}}$} & \multicolumn{1}{l}{$\rm{Ours}_{\rm{50K}}$} & \multicolumn{1}{l}{$\rm{TriVol}_{\rm{50K}}$} \\ \hline
			PSNR $\left( \uparrow \right)$  & 23.89 & 27.01 & 28.43 & 27.50 \\
			SSIM $\left( \uparrow \right)$  & 0.874 & 0.902 & 0.933 & 0.929 \\
			LPIPS $\left( \downarrow \right)$ & 0.203 & 0.117 & 0.183 & 0.179
		\end{tabular}%
	}
\caption{Complete Views Rendering Comparisons on the ShapeNet dataset with the novel view synthesis. The subscripts indicate the number of points.}
\label{CompleteViewsTable}
\end{table}
\par
\subsection{Repairing} \label{RESec4Repairing}
Thanks to the ability of generalization of GPN, we could generate the missing parts of point clouds directly from the existing cloud. And we could also fine-tune the reconstructed results based on the captured frames. To evaluate repairing ability, we use two metrics:
\begin{itemize}
	\item \textbf{Chamfer Distance (CD).} We could use Chamfer Distance to measure the dissimilarity between 3D point clouds.
	\begin{equation*}
		CD\left ( P,Q \right )  =\sum_{p\in P} \min_{q \in Q}\left \| p-q \right \|^2 + \sum_{q\in Q} \min_{p \in P}\left \| p-q \right \|^2
	\end{equation*}
	where $\left \| \bullet  \right \| $ denotes the Euclidean norm in $\mathbb{R} ^3$.
	\item \textbf{Minimum matching distance (MMD)} is proposed to complement coverage as a metric that measures quality. For each point cloud in the reference set, the distance to its nearest neighbor in the generated set is compted and averaged.
	\begin{equation*}
		MMD\left ( S_g,S_r \right ) = \frac{1}{\left |S_r  \right | } \sum_{Y \in S_r}\min_{X \in S_g}CD\left ( X, Y \right )  
	\end{equation*}
	where CD means the Chamfer Distance between two clouds.
\end{itemize}
\par
\noindent \textbf{Up-sampling.} We could up-sample the low-density cloud directly using our GPN framework, and keep the color and geometry of high-density cloud consistent with the low-density cloud and the captured frames. As the Table.\ref{UpsamplingComparisonTable} shows, our GPN model gives results comparable to TriVol and Points2NeRF.
\begin{figure}[htbp]
	\centering
	\includegraphics[scale=0.2]{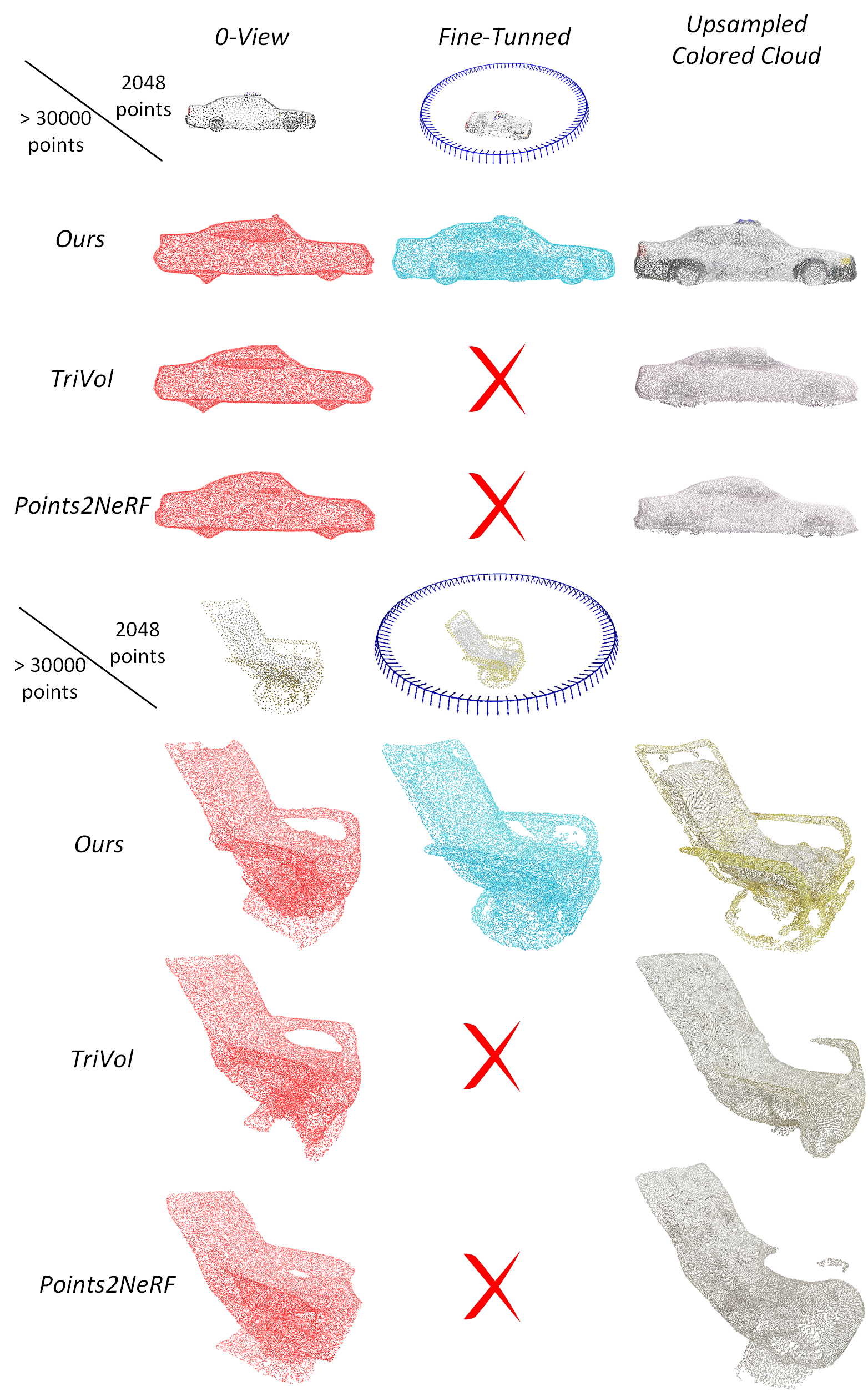}
	\caption{Up-sampling. Qualitative results of ours and other methods. There are 2048 points in the low-density cloud. We up-sampled it to 30000 points. Our GPN framework could generate the high-density colored cloud and achieve the best geometry results. Besides, we could optimize the color and geometry of clouds through fine-tunning on captured frames.}
	\label{USFig}
\end{figure}
\par
\begin{table}[htbp]
	\centering
	\resizebox{\columnwidth}{!}{%
		\begin{tabular}{l|ccc}
			Upsampling    & \multicolumn{1}{l}{Ours} & \multicolumn{1}{l}{TriVol\cite{TriVol}} & \multicolumn{1}{l}{Points2NeRF\cite{Points2NeRF}} \\ \hline
			car   & 12.18                    & 13.40                      & 14.88                           \\
			chair & 25.08                    & 25.14                      & 25.69                          
		\end{tabular}%
	}
	\caption{Point Cloud Upsampling results on ShapeNet compared using Chamfer Distance. The value of CD is computed on 16384 points and multiplied by $10^4$ }
	\label{UpsamplingComparisonTable}
\end{table}
\par
\noindent \textbf{Hole-filling.} When the captured frames could coverage the whole scene, but becaused of some unpredictable reasons, such as occluded area, un-sufficient light source. We could filled the missing parts of the reconstructed cloud through fine-tunning on the captured frames.
\par
\begin{figure}[htbp]
	\centering
	\includegraphics[scale=0.115]{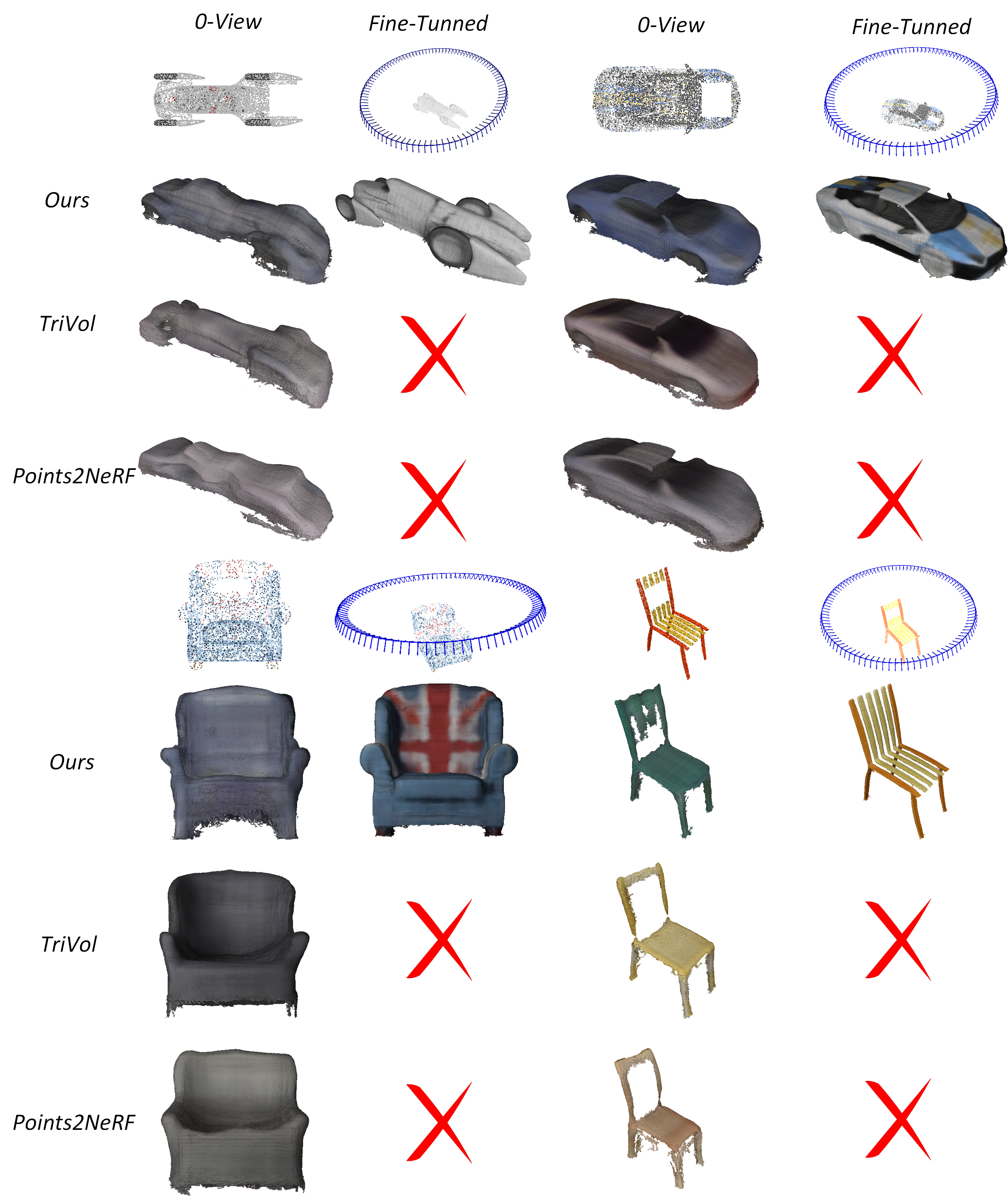}
	\caption{Hole-filling. Our results achieve better reconstruction than other methods, even in the 0-view tasks. We optimize the color and geometry of reconstructed cloud to make the result cloud high-density and high-resolution.}
	\label{HFFig}
\end{figure}
\par
\begin{table}[htbp]
	\centering
	\resizebox{\columnwidth}{!}{%
		\begin{tabular}{l|ccc}
			Hole-Filling   & \multicolumn{1}{l}{Ours} & \multicolumn{1}{l}{TriVol\cite{TriVol}} & \multicolumn{1}{l}{Points2NeRF\cite{Points2NeRF}} \\ \hline
			car   & 1.48                    & 17.01                      & 8.05                           \\
			chair & 6.74                    & 30.96                      & 22.01                          
		\end{tabular}%
	}
	\caption{Hole-Filling results for the GPN and other methods. MMD-CD scores are multiplied by $10^3$.}
	\label{HoleFillingTable}
\end{table}
\par
\subsection{Completion} \label{RESec4Completion}
When the catpured frame could not cover the whole scene. Based on our pre-trained GPN Completion framework, we could also complete the missing parts of whole scene. Our method could do more completion stuff than other classical point completion methods. We could even optimize the predicted cloud more consistent with the captuerd frames. It should be noted that we could complete the colored point cloud which most of other state-of-art completion method could only complete cloud without vertex colors.
\begin{figure}[htbp]
	\centering
	\includegraphics[scale=0.1]{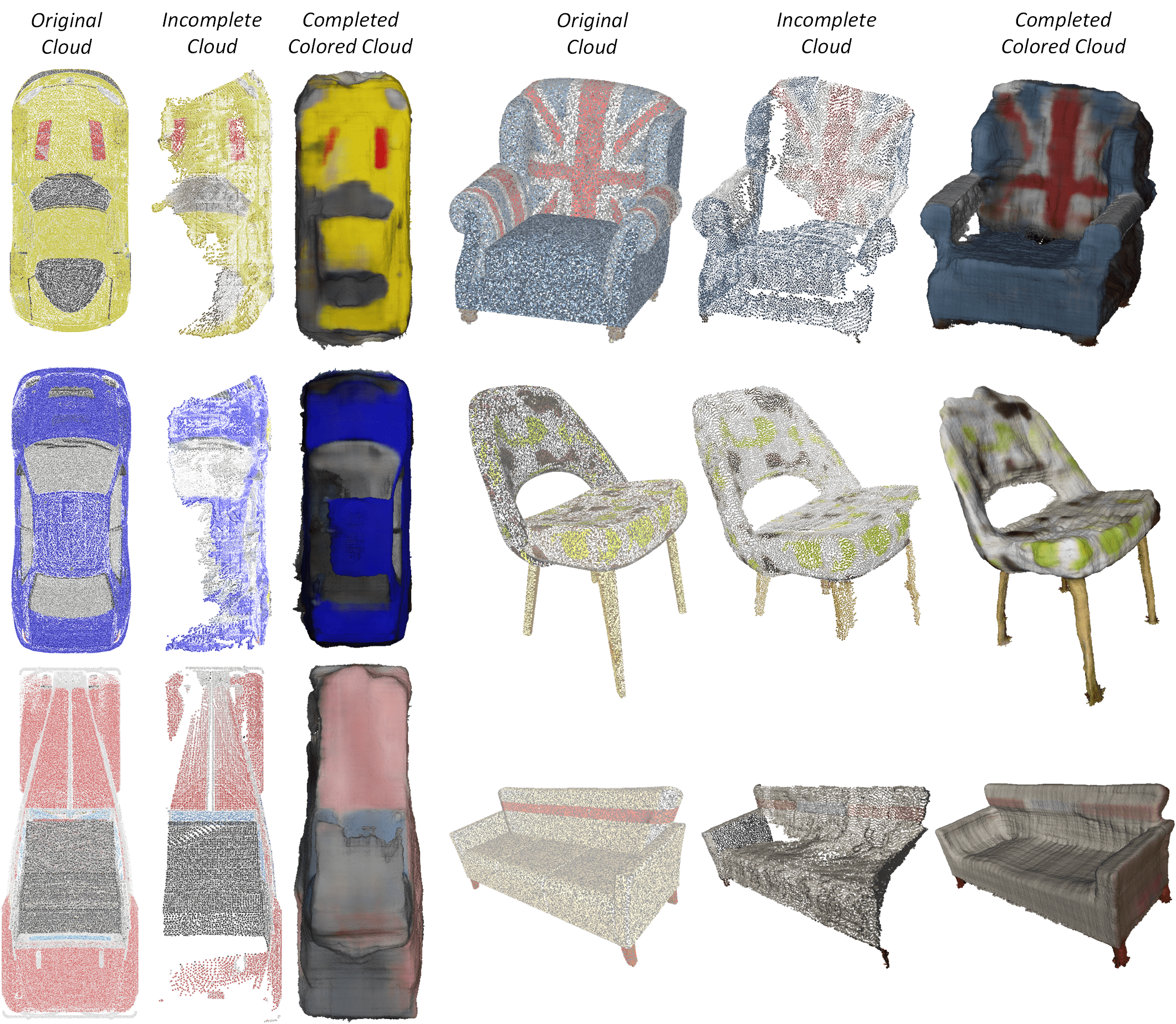}
	\caption{Completion Results. Qualitative completion results of different objects. We could notice that we could complete not only the geometry but also the vertex texture of completed mesh. As far as we know, we are the first work which could handle the color and geometry of completed cloud simultaneously.}
	\label{CFFig}
\end{figure}
\par
\subsection{GPU Memory} \label{GPUMemorySec}
The point cloud encoder in our GPN framework needs a uniform input of 2048 points. We do not need to provide a high-density point cloud as input for the point-based rendering section. Providing a high-density point cloud as input for the point-based rendering section is unnecessary because the hypernetwork-based NeRF network can generate an implicit NeRF scene using the latent code output obtained from the point-based encoder.
\par
\begin{table}[htbp]
	\centering
	\resizebox{\columnwidth}{!}{%
		\begin{tabular}{l|lll}
			Training		   & Ours                      & TriVol\cite{TriVol}                    & Points2NeRF\cite{Points2NeRF}               \\ \hline
			Memory (GB)        & \multicolumn{1}{c}{7.923} & \multicolumn{1}{c}{18.34} & \multicolumn{1}{c}{5.911}
		\end{tabular}%
	}
	\caption{GPU Memory in Training time.}
	\label{TrainingGPUTable}
\end{table}
\par
\begin{table}[htbp]
	\centering
	\resizebox{\columnwidth}{!}{%
		\begin{tabular}{l|ccc}
			Inference & \multicolumn{1}{l}{Ours} & \multicolumn{1}{l}{TriVol\cite{TriVol}} & \multicolumn{1}{l}{Points2NeRF\cite{Points2NeRF}} \\ \hline
			Memory (GB) & 5.321 & 11.684 & 3.821 \\
			PSNR $\left( \uparrow \right)$      & 18.59 & 18.56 & 12.30 \\
			SSIM $\left( \uparrow \right)$      & 0.739 & 0.707 & 0.519 \\
			LPIPS $\left( \downarrow \right)$    & 0.474 & 0.505 & 0.621
		\end{tabular}%
	}
	\caption{GPU Memory in Inference time.}
	\label{InferenceGPUTable}
\end{table}
\par
\section{Ablation Study} \label{AblationSec}
\noindent \textbf{The Effect of Rendering Backend in Inference Time.} In Sec.\ref{PNRSec}, we mentioned the reasons for choosing different Point-based NeRF renderers. We compare GPU memory and a comparison rendering results on ShapeNet after a 200-iter fine-tuning process in Table.\ref{RenderingBackendTable}.
\par
\begin{table}[htbp]
	\centering
	\resizebox{\columnwidth}{!}{%
		\begin{tabular}{l|cccc}
			Rendering Backend & \multicolumn{1}{l}{Memory (GB)} & \multicolumn{1}{l}{PSNR$\left( \uparrow \right)$} & \multicolumn{1}{l}{SSIM$\left( \uparrow \right)$} & \multicolumn{1}{l}{LPIPS$\left( \downarrow \right)$} \\ \hline
			Ours(NeRFAcc\cite{NeRFAcc})             & 5.321 & 18.59 & 0.739 & 0.474\\
			NeRF\cite{NeRF}       					& 7.438 & 12.63 & 0.536 & 0.658 \\
			Point-NeRF\cite{PointNeRF}  			& 5.695 & 17.98 & 0.685 & 0.539
		\end{tabular}%
	}
	\caption{Different Rendering Backend.}
	\label{RenderingBackendTable}
\end{table}
\par
\noindent \textbf{The Effect of Point Cloud Encoder in Training Time.} We explore the performance of different point encoders in transforming point cloud to latent code. We adopt a simplified PointNet-like point encoder. Following the comparison of different point cloud encoders in TriVol~\cite{TriVol}, the baselines are the encoders containing the point-based networks, including PointNet~\cite{PointNet}, PointNet++~\cite{PointNet++}, and the MinkowskiNet~\cite{MinkowskiNet}. The results are shown in Table.\ref{PointCloudEncoderTable}.
\par
\begin{table}[htbp]
	\centering
	\resizebox{\columnwidth}{!}{%
		\begin{tabular}{l|cccc}
			Point Encoder & \multicolumn{1}{l}{Ours} & \multicolumn{1}{l}{PointNet\cite{PointNet}} & \multicolumn{1}{l}{PointNet++\cite{PointNet++}} & \multicolumn{1}{l}{MinkowskiNet\cite{MinkowskiNet}} \\ \hline
			Memory (GB)             &        7.923                    &             9.264           & 9.598  & 11.911   \\
		\end{tabular}%
	}
	\caption{Different Point Cloud Encoder.}
	\label{PointCloudEncoderTable}
\end{table}
\par
\section{Conclusions} \label{ConclusionsSec}
We implement a generative point-based nerf framework, which could handle some basic point cloud process, such as novel object generation, complete the incomplete cloud based on the existing captured frames. We achieve a high-quality rendering and reconstruction results through TriVol and NeRFAcc. We could also remain the multi-view consistency of reconstructed cloud. In the continuous work, we could make GPN more fast and more accurate using gaussian splatting technology, and a more diversity generation ability using diffusion models.

\bibliographystyle{elsarticle-num}
\bibliography{cgi}
\end{sloppypar}
\end{document}